\documentclass[conference]{IEEEtran}
\IEEEoverridecommandlockouts
\usepackage{cite}
\usepackage{enumitem}
\usepackage{multirow}%
\usepackage{graphicx}
\usepackage{textcomp}
\usepackage{xcolor}
\usepackage{float}%
\usepackage{booktabs}%
\usepackage{makecell}
\usepackage{subcaption}
\usepackage{wrapfig}
\usepackage{bbding}
\usepackage{amsmath,amssymb,amsfonts}
\usepackage[ruled,linesnumbered]{algorithm2e}
\usepackage{soul}
\usepackage{url} 
\usepackage[hidelinks]{hyperref}
\usepackage{xspace}
\usepackage{pgfplots}
\usepackage{tikz}
\pgfplotsset{compat=1.17}
\usetikzlibrary{patterns}
\newcommand{\model}{UniVectorSQL}

\usepackage{array}

\def\BibTeX{{\rm B\kern-.05em{\sc i\kern-.025em b}\kern-.08em
    T\kern-.1667em\lower.7ex\hbox{E}\kern-.125emX}}
\begin{document}

\title{Text2VectorSQL: Towards a Unified Interface for Vector Search and SQL Queries}

\author{
\IEEEauthorblockN{Zhengren Wang\IEEEauthorrefmark{1},
Dongwen Yao\IEEEauthorrefmark{2},
Bozhou Li\IEEEauthorrefmark{1},
Dongsheng Ma\IEEEauthorrefmark{1},
Bo Li\IEEEauthorrefmark{1},\\
Zhiyu Li\IEEEauthorrefmark{4}\IEEEauthorrefmark{6},
Feiyu Xiong\IEEEauthorrefmark{4},
Bin Cui\IEEEauthorrefmark{1},
Linpeng Tang\IEEEauthorrefmark{3}\IEEEauthorrefmark{4},
Wentao Zhang\IEEEauthorrefmark{1}\IEEEauthorrefmark{3}\IEEEauthorrefmark{5}\IEEEauthorrefmark{6}}
\IEEEauthorblockA{
\IEEEauthorrefmark{1}Peking University
\IEEEauthorrefmark{2}Shanghai Jiao Tong University
\IEEEauthorrefmark{3}OriginHub Technology\\
\IEEEauthorrefmark{4}Institute for Advanced Algorithms Research, Shanghai
\IEEEauthorrefmark{5}Zhongguancun Academy
}
\IEEEauthorblockA{wzr@stu.pku.edu.cn, \{wentao.zhang, bin.cui\}@pku.edu.cn}
}

\maketitle
\begin{abstract}
\label{sec:abstract}
The proliferation of unstructured data poses a fundamental challenge to traditional database interfaces. While Text-to-SQL has democratized access to structured data, it remains incapable of interpreting semantic or multi-modal queries. Concurrently, vector search has emerged as the de facto standard for querying unstructured data, but its integration with SQL—termed VectorSQL—still relies on manual query crafting and lacks standardized evaluation methodologies, creating a significant gap between its potential and practical application.

To bridge this fundamental gap, we introduce and formalize \textbf{Text2VectorSQL}, a novel task to establish a unified natural language interface for seamlessly querying both structured and unstructured data. To catalyze research in this new domain, we present a comprehensive foundational ecosystem, including: (1) A scalable and robust pipeline for synthesizing high-quality Text-to-VectorSQL training data. (2) VectorSQLBench, the first large-scale, multi-faceted benchmark for this task, encompassing 12 distinct combinations across three database backends (SQLite, PostgreSQL, ClickHouse) and four data sources (BIRD, Spider, arXiv, Wikipedia). (3) Several novel evaluation metrics designed for more nuanced performance analysis. Extensive experiments not only confirm strong baseline performance with our trained models, but also reveal the recall degradation challenge: the integration of SQL filters with vector search can lead to more pronounced result omissions than in conventional filtered vector search. By defining the core task, delivering the essential data and evaluation infrastructure, and identifying key research challenges, our work lays the essential groundwork to build the next generation of unified and intelligent data interfaces. Our repository is available at \url{https://github.com/OpenDCAI/Text2VectorSQL}.
\renewcommand{\thefootnote}{}
\footnotetext{\IEEEauthorrefmark{6}Corresponding author.}
\end{abstract}


\section{Introduction}
\label{sec:introduction}
The modern data landscape is characterized by a fundamental dichotomy: the structured world of relational databases, navigated by the precise logic of SQL, and the vast unstructured data—estimated to constitute over 90\% of all enterprise data \cite{idc92}—increasingly queried through semantic vector search. This separation creates a critical barrier to unified data interfaces, and forces users to alternate different interfaces for SQL-based analytics and vector-based semantic search. 

On one hand, Text-to-SQL systems have made remarkable strides in translating natural language (NL) into structured queries, democratizing access to tabular data. However, they are inherently blind to the rich semantics embedded in unstructured content like text, images, and other modalities. Consequently, they cannot answer hybrid questions such as, ``Find scientific papers discussing learned index structures published after 2020".

Concurrently, vector search has emerged as the de facto standard for similarity-based retrieval, and its integration into relational databases—termed VectorSQL—promises to bridge this gap \cite{pgvector,myscale}. Yet, this integration remains at a nascent stage. Crafting effective VectorSQL queries is a manual, error-prone process demanding dual expertise in vector search and SQL queries. More critically, the academic  investigation to translate natural language into VectorSQL is entirely absent.

\begin{figure*}[t]
    \centering
    \includegraphics[width=\linewidth]{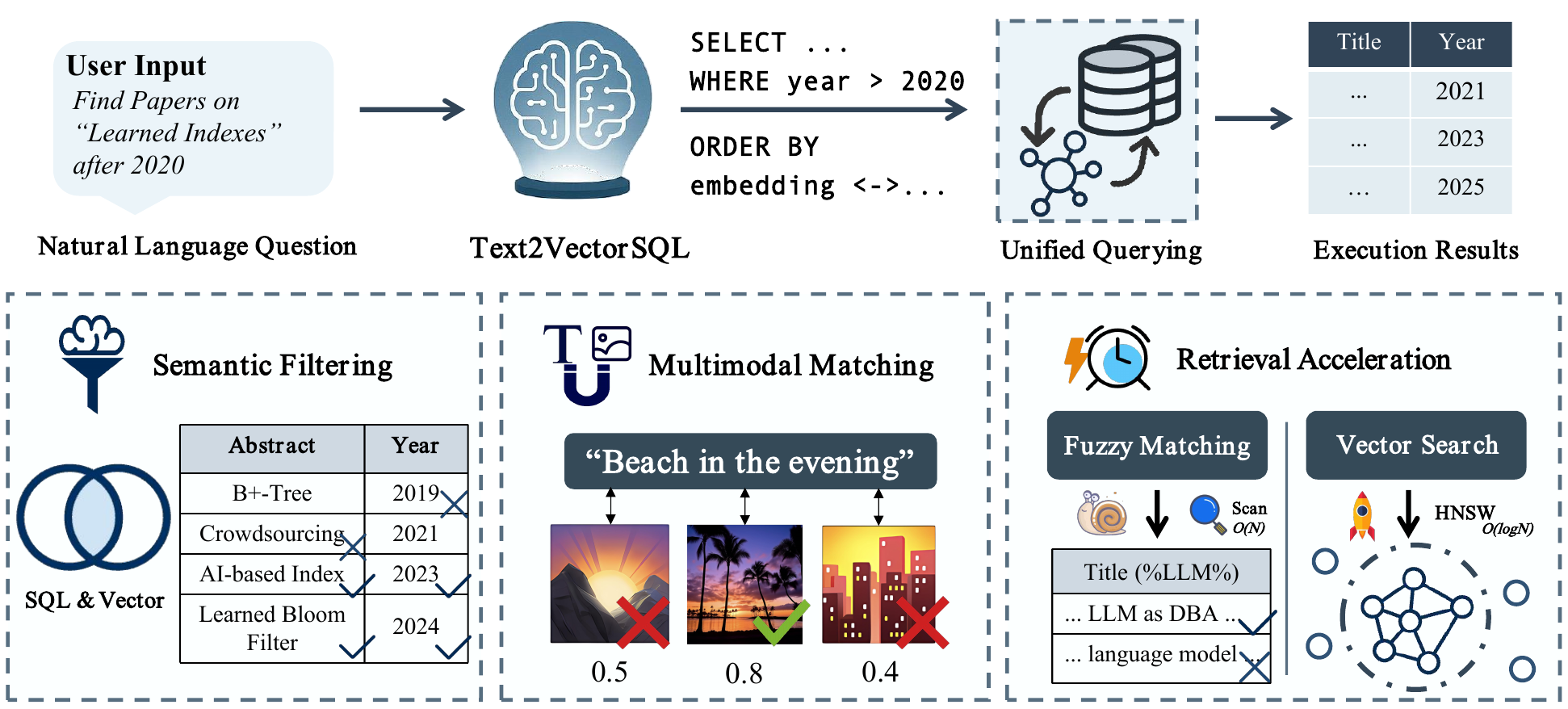}
    \caption{
    Illustration of the Text2VectorSQL task, with scenarios below showing how integrating SQL queries with vector search unlocks semantic filtering, multi-modal matching and retrieval acceleration. These capabilities are indispensable for universal natural language interfaces. 
    }
    \label{fig:task}
\end{figure*}

To bridge this fundamental gap and pave the way for a new class of data interaction paradigms, we propose Text2VectorSQL. However, realizing this vision is non-trivial and faces three core challenges: 
\begin{itemize}
    \item \textbf{An extreme scarcity of training data}, as no public datasets pairing natural language with VectorSQL queries exist for model traning or fine-tuning.
    \item \textbf{An evaluation void}, with no established benchmarks or metrics to measure the performance of Text2VectorSQL models, thus hindering reproducible research.
    \item \textbf{A tradeoff between accuracy and efficiency} in VectorSQL translation and execution, where the interplay between vector search and SQL filtering can lead to substantial but silent recall failures.
\end{itemize}

To address these foundational obstacles, we have built a foundational ecosystem. First, to overcome the data scarcity problem, we develop a scalable and robust pipeline for synthesizing high-quality, diverse Text-to-VectorSQL training data. Second, to fill the evaluation void, we construct VectorSQLBench, the first large-scale, multi-faceted benchmark for this task, spanning 12 distinct combinations of database backends (SQLite, PostgreSQL, ClickHouse) and data sources (BIRD, Spider, arXiv, Wikipedia). VectorSQLBench is accompanied by a suite of novel evaluation metrics, such as rank-based score and decomposed accuracy, for more comprehensive and fine-grained evaluation beyond execution accuracy.

Finally, leveraging this ecosystem, our extensive experiments yield crucial insights. While our trained models establish state-of-the-art performance, 
our analysis uncovers a critical and previously under-examined challenge we term \textbf{recall degradation}. We find that the popular ``post-filtering'' approach—where vector search results are subsequently filtered by SQL predicates—can lead to a more pronounced omission of correct results than in traditional filtered vector search. For instance, in complex queries involving `JOIN' operations, post-filtering can silently discard over 30\% of the correct results. This discovery elevates Text2VectorSQL from a mere translation problem to a complex challenge deeply intertwined with query optimization and execution strategy.

Our main contributions are summarized as follows:
\begin{itemize}
\item \textbf{Task Formalization.} We are the first to formally define the Text2VectorSQL task, establishing a clear objective for unifying natural language interaction with both structured and unstructured data, thereby overcoming the inherent limitations of traditional Text-to-SQL.

\item \textbf{A Foundational Ecosystem.} 
We provide a comprehensive ecosystem to catalyze research in this emerging area. This includes a scalable pipeline for synthesizing high-quality training data and VectorSQLBench, the first comprehensive benchmark for Text2VectorSQL evaluation. VectorSQLBench’s breadth (12 dataset combinations) and depth (proposed metrics) provide a robust foundation for holistic assessment.

\item \textbf{Empirical Analysis and Key Findings.} Through extensive experiments across multiple datasets and databases, we establish strong baseline performances for the task. More importantly, our analysis reveals the critical challenge of recall degradation in VectorSQL execution. This insight pinpoints a crucial direction for future research in co-optimizing query generation and execution.
\end{itemize}

Our data, models, and evaluation scripts are made publicly available at \url{https://github.com/OpenDCAI/Text2VectorSQL}.

\section{Preliminaries}
\label{sec:preliminaries}
In this section, we introduce foundational concepts that underpin our work: vector search and the VectorSQL paradigm.

\subsection{Vector Search}
Vector search retrieves items from a collection whose high-dimensional vector representations are most similar to a given query vector. This is the cornerstone of modern semantic and multi-modal information retrieval \cite{pan2024survey,pan2024vector,taipalus2024vector}.

\textbf{Definition 2.2 (Vector Search Task).} Let $\mathcal{D}$ be a collection of unstructured items (e.g., documents, images) and $\mathcal{E}: \mathcal{D} \rightarrow \mathbb{R}^d$ be an embedding model that maps any item $d \in \mathcal{D}$ to a $d$-dimensional vector $\vec{v} = \mathcal{E}(d)$. Given a query text $Q_{txt}$, its corresponding query vector $\vec{q} = \mathcal{E}(Q_{txt})$, a similarity function $\text{sim}(\cdot, \cdot)$ (e.g., cosine similarity or dot product), and an integer $k$, the vector search task is to find a result set $R_V \subset \mathcal{D}$ of size $k$ that contains the top-$k$ most similar items to the query:
$$R_V = \underset{R \subset \mathcal{D}, |R|=k}{\arg\max} \sum_{d_i \in R} \text{sim}(\vec{q}, \mathcal{E}(d_i))$$

For efficiency on large-scale datasets, this is often implemented using Approximate Nearest Neighbor (ANN) search algorithms for logarithmic time complexity. In many practical applications, this task is extended to filtered vector search, which requires retrieved items not only to be semantically similar to the query but also to satisfy some filter predicate.

\subsection{Text-to-VectorSQL}
\textbf{Definition 2.3 (VectorSQL Query).} A VectorSQL query, denoted as $Q_{VSQL}$, is conceptually composed of both structured filtering and semantic vector search components: 
\begin{itemize}
    \item A \textit{SQL Predicate Component} ($Q_S$): A standard SQL query or subquery that specifies constraints on the structured data columns (e.g., `WHERE' and `JOIN' clauses).
    \item A \textit{Vector Search Component} ($Q_V$): A semantic search directive, defined by a tuple $(\vec{q}, k)$, where $\vec{q} \in \mathbb{R}^d$ is the query vector and $k$ is the number of nearest neighbors.
\end{itemize}
The execution of a $Q_{VSQL}$ query yields a result set by coordinating $Q_S$ and $Q_V$. 
For filtered vector search, the simplified form of VectorSQL, this coordination can be categorized into two distinct strategies:
\begin{itemize}

    \item \textit{Pre-filtering (Filter-then-Search).} This strategy prioritizes accuracy by first using SQL predicates to define an exact subset of the data, and then performing vector search within that subset.
    
    \begin{enumerate}
        \item Filter: The SQL query $Q_S$ is executed to select a precise subset of items: $\mathcal{D}' = \{d \in \mathcal{D} \mid d \text{ satisfies } Q_S\}$.
        \item Search: A vector search is performed only on the filtered subset to find the top $k$ results: $R_{pre} = \text{ANN\_Search}(\vec{q}, k, \mathcal{D}')$.
    \end{enumerate}
    This method guarantees that all results adhere to the structured criteria but can be inefficient, as it may invalidate the assumptions of pre-built ANN indexes, potentially forcing a costly brute-force scan over $\mathcal{D}'$.

    \item \textit{Post-filtering (Search-then-Filter).} This strategy prioritizes speed by first performing an approximate vector search to narrow down the candidates, and then applying SQL filters.
    \begin{enumerate}
        \item Search: An ANN search is executed to retrieve an initial candidate set, $R_V = \text{ANN\_Search}(\vec{q}, k')$, where $k'$ could be larger than the final desired $k$ to account for items that will be filtered out.
        \item Filter: SQL predicates $Q_S$ are applied to retrieved candidates: $R_{post} = \{d \in R_V \mid d \text{ satisfies } Q_S\}$.
    \end{enumerate}
    Actually, SQL filters and semantic similarity are usually considered as hard and soft constraints respectively. While the execution of pre-filtering is regarded as high-fidelity, post-filtering is efficient but can suffer from recall degradation to retrieve enough correct results.
\end{itemize}

\begin{figure*}[tb]
    \centering
    \includegraphics[width=\linewidth]{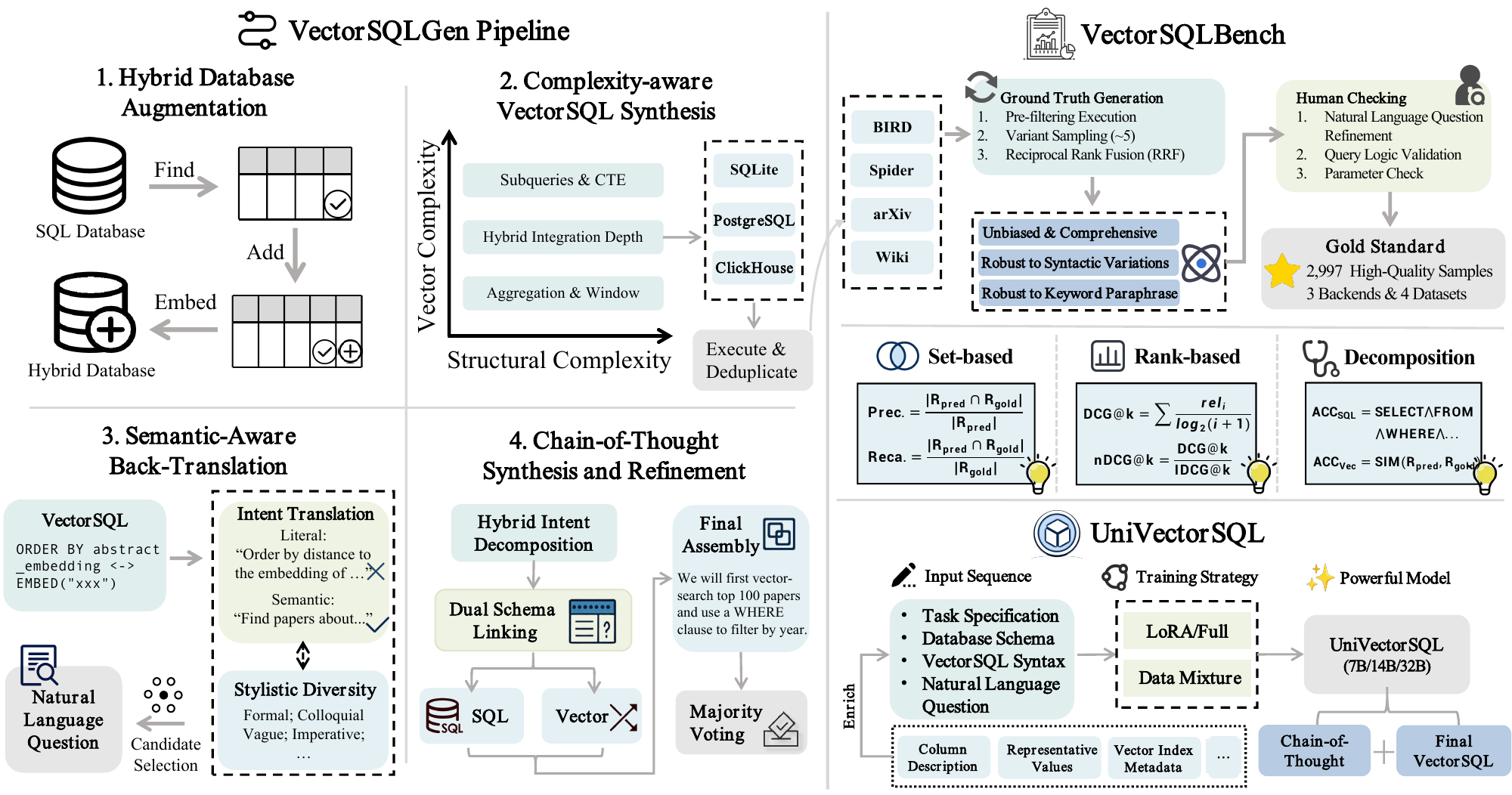}
    \caption{The Text2VectorSQL Ecosystem. The core component is \textbf{VectorSQLGen} pipeline, a large-scale, automated data synthesis engine that produces high-quality training samples. Then, the synthesized data is used to train our family of \textbf{UniVectorSQL} models. Concurrently, a curated subset of data undergoes a more rigorous, human-review process to create \textbf{VectorSQLBench}, our gold-standard evaluation benchmark with a suite of novel and fine-grained metrics.
    }
    \label{fig:ecosystem}
\end{figure*}

Finally, with the definition of VectorSQL established, we can define the Text-to-VectorSQL task.

\textbf{Definition 2.4 (Text-to-VectorSQL Task).} Given a hybrid database schema $S_{hybrid}$, which contains relational tables with vector columns, and a natural language question $Q_{NL}$, the Text-to-VectorSQL task is to learn a mapping function $f_{\text{T2VSQL}}$ that translates the question into a valid VectorSQL query $Q_{VSQL} = f_{\text{T2VSQL}}(Q_{NL}, S_{hybrid})$.

\section{The Foundational Ecosystem for Text2VectorSQL}\label{sec:method}
To systematically address the challenges of data scarcity, the absence of evaluation standards, and the lack of robust baselines in Text2VectorSQL, we have developed a foundational ecosystem as shown in Fig. \ref{fig:ecosystem}. The cornerstone of this ecosystem is \textbf{VectorSQLGen}, a scalable pipeline designed to synthesize high-quality training data. This pipeline fuels the development of our \textbf{UniVectorSQL} models. Concurrently, we constructed \textbf{VectorSQLBench}, a principled benchmark for holistic evaluation. 

\subsection{VectorSQLGen: A Synthetic Pipeline for Text2VectorSQL Training Data}

Addressing the extreme scarcity of training data is the primary prerequisite for progress in Text2VectorSQL. Inspired by the success of data synthesis paradigms like OmniSQL \cite{li2025omnisql}, we introduce VectorSQLGen, a novel pipeline that significantly extends this approach for Text2VectorSQL task. VectorSQLGen automates the creation of high-quality training quadruplets: \textless Hybrid Database, Natural Language Question, VectorSQL Query, Chain-of-Thoughts\textgreater. The pipeline operates through a carefully designed four-stage process.

\subsubsection{Hybrid Database Augmentation}

The first challenge is that standard SQL databases are purely relational. To synthesize meaningful VectorSQLs, we must enrich these databases with a semantic layer:

\begin{itemize}
    \item \textbf{Base Database Sourcing}: Real-world databases are scarce on the internet because enterprise databases often contain sensitive information. Despite this, the tabular data is abundant and also reflects real-world scenarios. To ensure diverse sources and enhance model generalization, we synthesize entirely new databases using web table corpora, following the methodology of OmniSQL \cite{omnisql}.

    \item \textbf{Target Column Identification}: We employ a LLM as the ``schema analyst" to autonomously identify columns that are semantically rich and suitable for vectorization. The LLM is prompted to analyze schema definitions (\texttt{CREATE TABLE} statements) and table contents (e.g., sample rows) to find columns containing free-form text, such as \texttt{abstracts} in a \texttt{papers} table or \texttt{descriptions} in a \texttt{products} table.

    \item \textbf{Semantic Content Augmentation}: For tables lacking rich textual content, the LLM is instructed to generate plausible unstructured text conditioned on the existing structured data in each row. For example, given a \texttt{papers} table row with columns \texttt{title}, \texttt{authors}, and \texttt{conference}, the LLM can author a concise \texttt{paper\_abstract}. This step augments semantic content which is contextually coherent with relational data.

    \item \textbf{Embedding and Schema Modification}: We employ some popular embedding models (e.g., all-MiniLM-L6-v2) to convert the content of identified or augmented columns into dense vector representations. Subsequently, the database schema is programmatically altered by adding new vector-type columns. The generated vectors are then inserted into new columns, transforming the relational database into a hybrid database ($S_{hybrid}$) capable of supporting both SQL and vector operations.
\end{itemize}

\subsubsection{Complexity-Aware VectorSQL Query Synthesis}
\label{subsubsec:step2_synthesis}

This stage is the core of our pipeline, designed to generate a diverse corpus of VectorSQL queries that are not only syntactically correct across multiple database backends but also span a wide spectrum of logical complexity. We move beyond a one-dimensional view of complexity to a more nuanced, two-dimensional framework.

\begin{figure*}[tb]
    \centering
    \includegraphics[width=\linewidth]{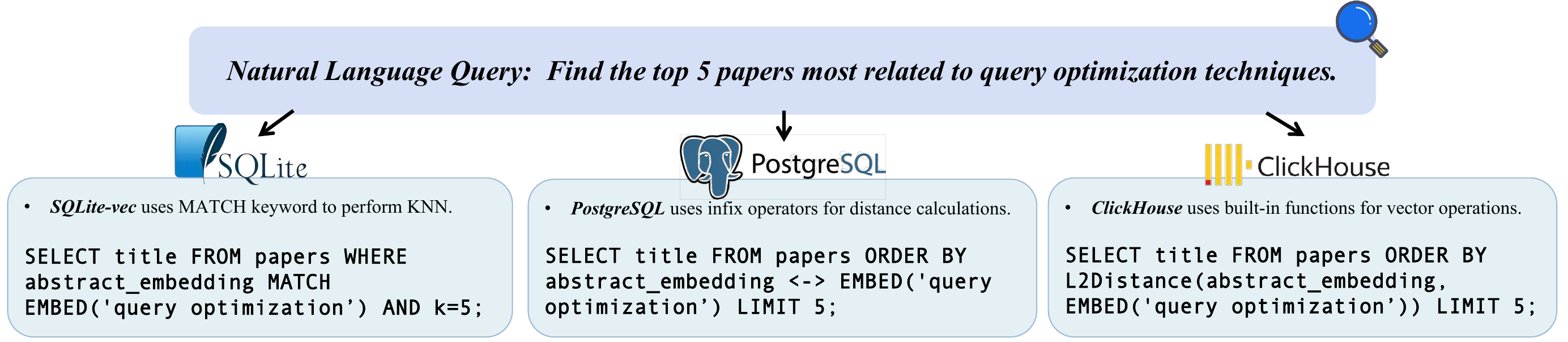}
    \caption{ Diverse syntax of vector operations in SQLite (via sqlite-vec), PostgreSQL (via pgvector) and ClickHouse.
    SQLite-vec generates distance column implicitly using virtual table mechanism and vec0 engine. PostgreSQL and ClickHouse support building indexes for approximate nearest neighbor (ANN) search, while SQLite-vec does not so far.
    }
    \label{fig:backend}
\end{figure*}

\begin{itemize}
    \item \textbf{Multi-Backend Syntax Adaptation}: A key practical challenge is the syntactic divergence of vector operations across database systems. Our benchmark incorporates SQLite (via sqlite-vec) \footnote{\url{https://github.com/asg017/sqlite-vec}}, PostgreSQL (via pgvector) \footnote{\url{https://github.com/pgvector/pgvector}}, and ClickHouse. To handle this, we designed backend-specific syntax templates. During query generation, the appropriate template is injected into the LLM prompt, guiding it to produce a syntactically valid query for the target backend. E.g., using \texttt{L2Distance} function in ClickHouse versus the \texttt{<->} operator in PostgreSQL. We provide a concrete example in Figure \ref{fig:backend}.

    \item \textbf{A Two-Dimensional Complexity Framework}: To ensure the synthesized data covers a wide spectrum of challenges, we define query complexity along two orthogonal axes: \textbf{Structural Complexity} and \textbf{Vectorial Complexity}.

    \begin{enumerate}
        \item \textbf{Structural Complexity}: This axis captures the difficulty of the traditional SQL portion of the query, covering both basic data retrieval and advanced data analysis. We control for metrics such as:
        \begin{itemize}
            \item \emph{Subquery \& CTE Depth}: The level of nesting for subqueries and the number of \texttt{WITH} clauses.
            \item \emph{Join Complexity}: The number of involved tables and type (\texttt{INNER}, \texttt{LEFT}, etc.) of \texttt{JOIN} operations.
            \item \emph{Aggregation, Grouping \& Set Operations}: The presence of \texttt{GROUP BY}, \texttt{HAVING}, and window functions, as well as the use of set functions like \texttt{UNION}, \texttt{INTERSECT}, or \texttt{EXCEPT}.
        \end{itemize}

        \item \textbf{Vectorial Complexity}: This novel axis quantifies the difficulty inherent in the semantic search component of the query. It is composed of three sub-dimensions:
        
        \begin{itemize}
            \item \textit{Number of Vector Operations}: Queries can range from a single vector search to multiple vector searches, which may require joining or intersecting results from different semantic queries (e.g., ``Find papers with a methodology related to `B-tree' and an application domain similar to `Spanner'.").

            \item \textit{Query Intent Complexity:} This evaluates the cognitive leap and world knowledge required to translate the user's natural language phrasing into precise query keywords. We define a three-tier hierarchy: (1) \textit{Entity-level:} This is the most straightforward case where the query target is a direct, explicit entity such as `pgvector'; (2) \textit{Concept-level:} The user describes an abstract concept, category, or topic rather than a specific entity, such as `extension for vector search in relational database'; (3) \textit{Reasoning-level:} The intent is highly abstract, relational, or requires reasoning for Text2VectorSQL models to formulate and embedding models to understand, such as `database with both SQL interface and vector search capability but different storage format from ClickHouse'.
            
            \item \textit{Hybrid Integration Depth}: This crucial metric measures how tightly the vector search is coupled with SQL predicates---a key factor influencing the \textbf{recall degradation} problem. From simple to complex, we define three levels: (1) \textit{Non-Integration}: Vector search is performed independently (e.g., in a CTE or subquery), and its results (e.g., primary keys) are used to filter the main query via \texttt{WHERE IN} or a \texttt{JOIN}; (2) \textit{WHERE-Integration}: The vector search is performed over a subset of data already filtered by a \texttt{WHERE} clause; (3) \textit{JOIN-Integration}: The vector search is performed on a table after it has been joined with other tables, meaning the filtering conditions come from related tables. This is the most complex scenario and where post-filtering strategies are most likely to fail.
        \end{itemize}
    \end{enumerate}

    \item \textbf{Generation and Post-processing}: During synthesis, we randomly sample a target complexity from this 2D complexity space and prompt the LLM to generate a corresponding VectorSQL query. All generated queries are executed against the database to filter out syntactic errors, execution timeouts, or those that yield empty results. Finally, queries are templatized to remove string literal and then deduplicated to maximize diversity.
\end{itemize}

\subsubsection{Semantic-Aware Question Back-Translation}
\label{subsubsec:step3_backtranslation}

With a corpus of high-quality VectorSQL queries, we employ the back-translation technique \cite{edunov2018understanding} to generate corresponding natural language questions, i.e. VectorSQL-to-Question.

\begin{itemize}
    \item \textbf{Semantic Intent Translation}: The core of this step is explicitly prompting the LLM to understand and interpret the \emph{semantic intent} of the vector search component. For a clause like \texttt{ORDER BY abstract\_embedding <-> EMBED("query optimization")}, the prompt instructs the LLM to generate questions like ``Find papers about techniques for speeding up database queries", not a literal translation like ``Order papers by the distance of their abstract embedding to ...".

    \item \textbf{Stylistic Diversity}: We adopt the nine linguistic styles defined in OmniSQL (i.e., formal, colloquial, imperative, interrogative, descriptive, concise, vague, metaphorical, and conversational) to generate diverse questions for each VectorSQL. This ensures the training data reflects the variety of real-world user inputs, and enhances model robustness. For example, a formal query might be ``Retrieve top-10 articles related to database published after 2022," while a vague version could be ``Show me some recent DB papers from the last couple of years."

    \item \textbf{Semantic Consistency Selector:} For each VectorSQL, we generate multiple candidate questions. To automatically select the highest-quality one, we embed all candidates into a vector space and identify the question closest to the semantic centroid of the cluster, a technique proven to be effective for selecting the most representative sample and enhancing data quality \cite{omnisql,Rossiello2017@centroid}.
\end{itemize}

\subsubsection{Chain-of-Thought (CoT) Synthesis and Refinement}
\label{subsubsec:step4_cot}

The final stage serves a dual purpose: generating a detailed, step-by-step reasoning process to be used as a powerful training signal, and performing a final round of quality assurance through self-correction.

\begin{itemize}
    \item \textbf{Extended CoT Structure for Hybrid Queries}: We design a CoT structure tailored for Text2VectorSQL that forces the LLM to reason about both structured SQL filters and semantic vector search components:    
    \begin{enumerate}
        \item \emph{Hybrid Intent Decomposition}: The CoT begins by analyzing the NL question to separate structural constraints (e.g., ``published after 2020") from semantic search intents (e.g., ``discussing relational databases").
        \item \emph{Dual Schema Linking}: It then links the structural parts to specific tables and columns, and the semantic part to the appropriate vector column(s).
        \item \emph{Vector Query Formulation}: It explicitly describes how the semantic intent is translated into a vector search clause, involving the reasoning on query keywords.
        \item \emph{SQL Predicate Formulation}: It articulates the construction of the standard SQL clauses (\texttt{WHERE}, \texttt{JOIN}, etc.).
        \item \emph{Final Assembly and Strategy Explanation}: Crucially, the final step assembles the complete VectorSQL query and explicitly explains the chosen integration strategy. For instance, ``We will first perform a vector search for the top 100 papers and then use a WHERE clause to filter by year." This makes the model's reasoning process more transparent and controllable.
    \end{enumerate}

    \item \textbf{Self-Correction via Majority Voting}: To maximize the correctness of the final training samples, we generate multiple CoT solutions for each question. We then extract the VectorSQL query from each CoT, execute them, and compare their execution results. The CoT whose query result has the largest intersection with the majority vote result is selected as the canonical, most reliable training instance. This refinement corrects subtle errors and sub-optimal choices made in the initial query synthesis. %
\end{itemize}

To build a Text2VectorSQL model for multiple database backends, we collect about 8k training samples each backend, and aggregate them into SynVecSQL-24k as our training set.

\begin{table*}[t]
\centering
\caption{Comprehensive Statistics and Complexity Distribution of VectorSQLBench. This table presents the overall scale of the benchmark with a detailed breakdown of query complexity. Structural complexity is rated based on SQL predicate difficulty (E: Easy, M: Medium, H: Hard, EH: Extra Hard). Vectorial complexity is focused on the integration depth of the vector search with SQL predicates (Non: Non-Integrate, WHERE: WHERE-Integrate, JOIN: JOIN-Integrate). We also supplement the statistics of our synthetic training data (SynVecSQL-24k) for reference.}
\label{tab:benchmark_stats}
\renewcommand{\arraystretch}{1.1}
\resizebox{\textwidth}{!}{%
\begin{tabular}{l ccccc c cccc ccc}
\toprule
\multirow{2.5}{*}{\textbf{Dataset}} & \multicolumn{5}{c}{\textbf{Data Scale}} & \multirow{2.5}{*}{\textbf{\# Samples}} & \multicolumn{4}{c}{\textbf{Structural Complexity Breakdown}} & \multicolumn{3}{c}{\textbf{Vectorial Complexity Breakdown}} \\
\cmidrule(lr){2-6} \cmidrule(lr){8-11} \cmidrule(lr){12-14}
& \textbf{\# DBs} & \textbf{\# Tables} & \textbf{\# Cols} & \textbf{\# Vec. Cols} & \textbf{\# Total Rows} & & \textbf{E} & \textbf{M} & \textbf{H} & \textbf{EH} & \textbf{Non} & \textbf{WHERE} & \textbf{JOIN} \\
\midrule
BIRD-Vec & 69 & 519 & 3,805 & 140 & 362,892,413 & 1,278 & 438 & 315 & 264 & 261 & 441 & 400 & 437 \\
Spider-Vec & 166 & 872 & 5,955 & 497 & 1,604,329 & 1,080 & 351 & 276 & 228 & 225 & 317 & 374 & 389 \\        
arXiv-Vec & 1 & 7 & 29 & 3 & 28,475,592 & 261 & 69 & 99 & 39 & 54 & 95 & 92 & 74 \\          
Wikipedia-Vec & 1 & 5 & 35 & 6 & 7,973 & 378 & 84 & 96 & 93 & 105 & 116 & 139 & 123 \\        
\textbf{Total} & \textbf{237} & \textbf{1,403} & \textbf{9,824} & \textbf{646} & \textbf{392,980,307} & \textbf{2,997} & \textbf{942} & \textbf{786} & \textbf{624} & \textbf{645} & \textbf{969} & \textbf{1,005} & \textbf{1,023} \\
\midrule
Training Data & 1,839 & 15,586 & 117,884 & 7,467 & 20,270 & 24,421 & 9,934 & 4,324 & 6,142 & 4,021 & 7,078 & 8,092 & 9,251 \\
\bottomrule
\end{tabular}%
}
\end{table*}

\subsection{VectorSQLBench: A Principled Benchmark for Holistic Text2VectorSQL Evaluation}
While VectorSQLGen addresses the data scarcity challenge for model training, the field still lacks a standardized, high-quality benchmark. This void makes it difficult to compare different approaches, diagnose model weaknesses, and measure true progress. To fill this gap, we introduce \textbf{VectorSQLBench}, a multi-faceted and cross-backend benchmark with rigorous annotation. Beyond end-to-end execution accuracy, it is designed to provide fine-grained, diagnostic insights into a model's capabilities across a spectrum of complexities. To construct a comprehensive testbed, we selected base data sources that cover both complex structured schemas (from the classic \textbf{BIRD} \cite{bird} and \textbf{Spider} \cite{spider} benchmarks) and rich unstructured content (from \textbf{arXiv} papers \footnote{\url{https://www.kaggle.com/datasets/Cornell-University/arxiv}} and \textbf{Wikipedia} articles \footnote{\url{https://www.kaggle.com/datasets/jacksoncrow/extended-wikipedia-multimodal-dataset}}). The arXiv dataset contains unstructured titles, abstracts and comments, whereas the Wikipedia dataset offers images, captions, paragraphs, headings and so on.

\subsubsection{The Construction Pipeline of VectorSQLBench}
The construction of VectorSQLBench incorporates both automatic data synthesis and human expert review, which follows a meticulous three-stage pipeline designed to ensure diversity, correctness, and naturalness.
\begin{itemize}
    \item \textbf{Candidate Seeding via Stratified Sampling.} To create a representative and challenging benchmark, we employ stratified sampling based on our two-dimensional complexity framework (see Section \ref{subsubsec:step2_synthesis}). By sampling instances from each cell of the \textit{Structural Complexity} × \textit{Vectorial Complexity} matrix, we ensure that VectorSQLBench ranges from simple, single-table semantic searches to highly complex queries that require joining multiple tables before performing a vector search, thereby providing a comprehensive testbed for model capabilities.

    \item \textbf{Principled Ground-Truth Generation.} For an evaluation benchmark, correctness and completeness are paramount. The post-filtering strategy, while efficient, is unacceptable for establishing ground truth due to its inherent recall degradation problem. Therefore, for every sampled candidate, we collect a \textbf{golden result set} ($R_{gold}$) by mandating a \textbf{pre-filtering} execution strategy. Furthermore, to account for the possibility of multiple valid VectorSQL formulations for a single question, our pipeline samples five VectorSQL variants for each question. We execute all variants using the pre-filtering strategy and define $R_{gold}$ as the sorted union of execution results with identical columns via reciprocal rank fusion (RRF) \cite{cormack2009reciprocal}. This process establishes an unbiased and comprehensive ground truth that is resilient to syntactic variations or keyword paraphrasing \footnote{For VectorSQL in our paper, we regard the results of pre-filtering execution as the ground-truth. However, the execution or explicit declaration of post-filtering can be efficient and acceptable for vague queries.}.

\item \textbf{Expert Annotation and Refinement.} Human review is the final stage that elevates the benchmark to a gold standard. Each sampled instance undergoes a thorough review by annotators with expertise in both database systems and natural language processing. The review process mainly involves three tasks:
\begin{enumerate}
    \item \textit{Natural Language Question Refinement}: Annotators assess the \textit{naturalness} and \textit{clarity} of the back-translated question. Ambiguous or stilted questions are paraphrased to better reflect real-world user intent and tone.
    \item \textit{Query Logic Validation}: Experts validate that the associated VectorSQL is a direct and logical translation of the refined natural language question. This includes correcting subtle logical flaws and ensuring executability. Any sample that fails to execute is discarded.
    \item \textit{Parameter Check}: A critical step is to check query parameters that depend on user intent. For example, based on the phrasing (e.g., ``find the most similar paper'' vs. ``show me some related papers''), experts annotate the most appropriate value for the top-$k$ parameter, providing a reference value for evaluating a model's ability to infer such nuances.
\end{enumerate}
\end{itemize}

\subsubsection{Benchmark Composition and Statistics}
The principled pipeline yields a comprehensive benchmark, VectorSQLBench, which is diverse in its data sources and challenging in its query complexity. Table \ref{tab:benchmark_stats} provides a detailed overview of the benchmark. It comprises a substantial 2,997 high-quality samples  distributed across 3 database backends and 4 distinct data sources (BIRD-Vec, Spider-Vec, arXiv-Vec, and Wikipedia-Vec), covering both complex structured schemas and rich unstructured content.

Notably, the proportion of high-complexity queries is significant, and the benchmark is balanced across two complexity dimensions. The axis of structural complexity, rating SQL predicate difficulty, includes 942 Easy, 786 Medium, 624 Hard, and 645 Extra Hard samples. The axis of vectorial complexity, focuses on the integration depth of the vector search, contains 969 Non-Integrate samples, 1,005 WHERE-Integrate samples, and 1,023 JOIN-Integrate samples in total. The high number of queries in the Hard/Extra Hard (1,269) and JOIN-Integration (1,023) categories presents a significant challenge for current and future Text2VectorSQL models.

\begin{figure}[tb]
    \centering
    \includegraphics[width=\linewidth,clip,trim=20mm 0mm 20mm 0mm]{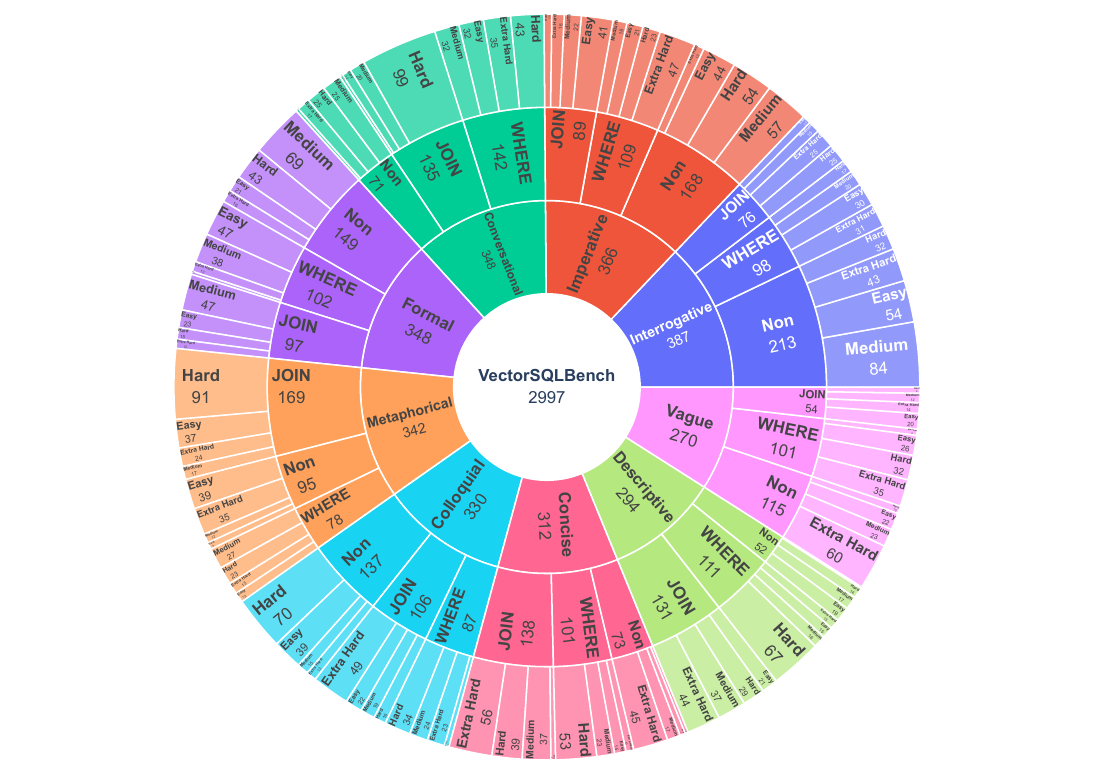}
    \caption{Diversity of the VectorSQLBench. The hierarchical structure breaks down the dataset across three dimensions: (1) nine distinct linguistic styles (inner ring), ensuring robustness to varied user phrasing; (2) three levels of vectorial complexity (middle ring), detailing how vector search is integrated with SQL (Non, WHERE, JOIN); and (3) four levels of structural SQL difficulty (outer ring), ranging from Easy to Extra Hard. This multi-faceted design benefits comprehensive coverage of the challenges in Text2VectorSQL.}
    \label{fig:sun_graph}
\end{figure}

Figure \ref{fig:sun_graph} further reveals the multi-layered design intended to stress-test models across a wide range of scenarios. The first layer (inner ring) demonstrates broad linguistic diversity, distributing the 2,997 queries across nine distinct linguistic styles to reflect the wide spectrum of real-world user expression. These styles include, for instance, Formal (348 samples), Colloquial (330 samples), Vague (270 samples), and Descriptive (294 samples). The subsequent layers dissect query complexity-the vectorial complexity and structural SQL complexity respectively. This hierarchical distribution guarantees VectorSQLBench as a rigorous evaluation suite.

\subsubsection{A Suite of Tailored Evaluation Metrics}
\label{metrics}
To facilitate a practical and insightful analysis, we move beyond Execution Accuracy (EX) \cite{bird}, which is ill-suited for Text2VectorSQL. The reliance of EX on exact matching of execution results is brittle for vector search where approximate results are essentially unavoidable and acceptable. Therefore, we propose a suite of more robust and diagnostic metrics.

\begin{itemize}
\item \textbf{Set-based Score}: This is our basic metric for end-to-end performance. It measures the overlap between the result set of predicted VectorSQL ($R_{\text{pred}}$) and the gold result set ($R_{\text{gold}}$). It is robust to capture semantic equivalence and tolerate syntactic variations. The score is calculated as follows, where F1 score is calculated as the harmonic mean of precision and recall.
$$ \text{Precision} = \frac{|R_{\text{pred}} \cap R_{\text{gold}}|}{|R_{\text{pred}}|}, \quad \text{Recall} = \frac{|R_{\text{pred}} \cap R_{\text{gold}}|}{|R_{\text{gold}}|} $$

\item \textbf{Rank-based Score}: For queries where the order of semantic relevance is important (e.g., ``Find the top 5 most similar articles"), set-based metrics are insufficient. We therefore also employ Normalized Discounted Cumulative Gain (nDCG@k) \cite{jarvelin2002cumulated}, a standard information retrieval metric. It rewards models for placing more relevant items higher in the result list. Given a predicted ranking, the nDCG@k is calculated as:
$$\text{DCG}@k = \sum_{i=1}^{k} \frac{\text{rel}_i}{\log_2(i+1)}, \quad \text{nDCG}@k = \frac{\text{DCG}@k}{\text{IDCG}@k}$$
where $\text{rel}_i$ is the relevance of the item at rank $i$ (1 if in $R_{gold}$, 0 otherwise), and IDCG is the DCG score of the ideal ranking. Other information retrieval metrics such as MRR and MAP are also applicable \cite{zhao2024retrieval,wang_qaencoder_2024}.

\item \textbf{Decomposed Accuracy}: To offer deeper diagnostic insights into why a model fails, we decompose the evaluation into two orthogonal components: the correctness of the SQL skeleton and the vector search part.
\begin{itemize}
    \item \textit{SQL Skeleton Accuracy ($ACC_{SQL}$)}: This metric assesses the structural correctness of the non-vector portion of the query. We use an LLM-based evaluator to parse the query and individually verify the correctness of its core clauses: \texttt{FROM} (including tables and JOIN conditions), \texttt{SELECT} (examining columns and aggregations), \texttt{WHERE} (non-vector conditions), \texttt{GROUP BY}, and \texttt{ORDER BY} (non-vector sorting). A query is awarded an $ACC_{SQL}$ of 1 if and only if all its structural components are deemed correct. %
    
    \item \textit{Vector Component Accuracy ($ACC_{Vec}$)}: This metric isolates and measures the ability to correctly interpret user's unstructured semantics. This metric should be zero if predicted queries select vector columns incorrectly. Otherwise, we assess the quality of the query keywords by computing the cosine similarity between the embeddings of predicted keywords and ground-truth keywords derived from expert annotation (i.e., the BertScore \cite{zhang2019bertscore}). This robustly handles variations in phrasing (e.g., ``AI research" vs. ``papers on artificial intelligence").
\end{itemize}

\end{itemize}

\subsection{UniVectorSQL: A Family of State-of-the-Art Open-Source Text-to-VectorSQL LLMs}\label{sec:UniVectorSQL}
Leveraging the synthetic data from VectorSQLGen, we introduce \textbf{UniVectorSQL}, a family of powerful, open-source Text2VectorSQL models at 7B, 14B, and 32B scales. In this section, we describe the format of input-output sequence and outline the objective and strategy for model training.

\subsubsection{Input-Output Construction}
The input sequence combines the task definition, the database schema, the VectorSQL syntax and notes, and a natural language question. Following prior work~\cite{Rajkumar2022@eval-llm-text2sql, Li2024@codes, Talaei2024@chess, omnisql}, we format the schema as \texttt{CREATE TABLE} statements and enrich them with the following elements as comments: (1) \textit{Column Descriptions}: to assist the LLM in identifying the correct columns referenced in the question, particularly when column names are ambiguous (e.g., abbreviations); (2) \textit{Representative Values}: to inform the LLM about the format of stored values, aiding in the use of advanced functions like \texttt{STRFTIME}; (3) \textit{Question-Relevant Values}: to help the LLM to generate accurate predicates, especially when the question mentions specific database values; (4) \textit{Vector Metadata}: to teach the LLM the content and origin of embedding columns (e.g., ``A 384-dim semantic vector for the paper abstract from all-MiniLM-L6-v2 encoder''), enhancing understanding of embedding columns and achieving automatic selection of embedding models. Note that we incorporate VectorSQL syntax in the prompts, as well as well-designed notes including few-shot examples, so that small LLMs can handle Text2VectorSQL task with in-context learning capabilities.

The output sequence is the vector-aware CoT with final VectorSQL, which is generated by the VectorSQLGen pipeline. 

\subsubsection{Training Objective and Strategy}
The UniVectorSQL models are developed by fine-tuning open-source models. We employ the standard autoregressive language modeling objective, where the model learns to predict the next token in a sequence conditioned on the preceding tokens.
\begin{equation}
\text{Loss} = - \mathbb{E}_{(x, y) \sim D} \left[\textstyle \sum_{t} \log P_{\theta}(y_{t} \mid x, y_{<t}) \right],
\end{equation}
Here, $x$ is the aforementioned input sequence, and $y$ is the target output, our specialized vector-aware CoT. To ensure UniVectorSQL excels at hybrid queries without sacrificing proficiency in complex, standard SQL, we also consider to fine-tune the models on a curated mixture of our synthetic Text2VectorSQL corpus and established Text2SQL datasets like SynSQL-2.5M \cite{omnisql}. This blended training strategy ensures the model acquires new vector search capabilities while retaining its mastery over traditional SQL generation. We leave implementation details in Section~\ref{sec:impl_details} for interested readers.

\section{Experiments}
\label{sec:experiments}
In this section, we conduct extensive experiments to validate the effectiveness of our proposed ecosystem and identify challenges for future research.

\subsection{Experimental Setup}

\subsubsection{Benchmarks and Metrics}
Our main experiments are conducted on \textbf{VectorSQLBench}, the comprehensive benchmark built from four diverse data sources: BIRD, Spider, arXiv, and Wikipedia. To ensure robustness and generality, we evaluate all models across three distinct database backends that feature different VectorSQL syntax: SQLite (with sqlite-vec), PostgreSQL (with pgvector), and ClickHouse. We employ a suite of metrics for a holistic evaluation: rank-based score (nDCG@10), set-based score (F1 score) and decomposed accuracy for diagnostic purposes. For training data mixing experiments, we also consider BRID-dev benchmark and the execution accuracy metric. 

\subsubsection{Baselines}
We compare our \model{} family with a range of powerful open-source and closed-source LLMs. (1) \textbf{Our Models}: \model-7B, \model-14B, \model-32B, as well as their LoRA fine-tuned versions; (2) \textbf{Closed-Source LLMs}: Claude-3/4, GPT-4, Gemini-2.5 and Grok3/4 series serve as reference points for state-of-the-art performance; (3) \textbf{Open-Source LLMs}: We include leading instruction-tuned and code-specialized models such as Meta-Llama-3.1 \cite{llama3}, Qwen2.5-Coder \cite{hui2024@qwen2.5-coder}, DeepSeek-Coder \cite{Guo2024@deepseek-coder} and OmniSQL \cite{omnisql}. This open-source category is further broken down into 7B-8B models, 14B-22B models, and models larger than 32B like DeepSeek-V3.1 \cite{deepseek2024@deepseek-v3}.

\begin{table*}[t]
    \centering
    \caption{Performance evaluation (nDCG@10) and execution success rate (in \%) of various LLMs on VectorSQLBench across three database backends. Our \model{} models, fine-tuned on our synthetic SynVecSQL-24K, achieve state-of-the-art performance among open-source models and are competitive with top proprietary APIs. Note that, the input prompts are the same across models for fairness. Best results in each category are in \textbf{bold}. ``DSC'' is the abbreviation of ``DeepSeek-Coder''.}
    \label{tab:backend_results}
    \setlength{\tabcolsep}{3pt}
    \renewcommand{\arraystretch}{1.1}
    \resizebox{\textwidth}{!}{
    \begin{tabular}{l|cccc|cccc|cccc|c}
        \toprule
        \multirow{2.5}{*}{\textbf{LLM}} & \multicolumn{4}{c|}{\textbf{SQLite}} & \multicolumn{4}{c|}{\textbf{PostgreSQL}} & \multicolumn{4}{c|}{\textbf{ClickHouse}} & \multirow{2.5}{*}{\textbf{Average}} \\
        \cmidrule(lr){2-5} \cmidrule(lr){6-9} \cmidrule(lr){10-13}
        & \textbf{BIRD} & \textbf{Spider} & \textbf{arXiv} & \textbf{Wiki} & \textbf{BIRD} & \textbf{Spider} & \textbf{arXiv} & \textbf{Wiki} & \textbf{BIRD} & \textbf{Spider} & \textbf{arXiv} & \textbf{Wiki} & \\
        \midrule
        \multicolumn{14}{c}{\textbf{Closed-source LLMs}} \\
        \midrule
        Claude-3.5-Haiku      & 28.7 / 57.5 & 27.2 / 59.7 & 13.7 / 51.7 & 20.1 / 58.6 & 39.0 / 82.2 & 34.9 / 75.2 & 17.6 / 90.9 & 31.7 / 85.0 & 35.6 / 92.7 & 39.7 / 91.6 & 27.0 / 89.9 & 33.8 / 92.4 & 29.1 / 77.3 \\
        Claude-3.7-Sonnet      & 36.7 / 68.8 & 35.2 / 74.7 & 29.8 / 69.7 & 28.9 / 66.9 & 42.9 / 85.3 & 50.0 / 90.3 & 35.9 / 89.1 & 36.5 / 91.7 & 42.5 / 96.4 & \textbf{47.3 / 94.1} & 39.7 / 94.9 & 36.9 / 94.7 & 38.5 / 84.7 \\
        Claude-4-Sonnet      & 37.9 / 70.9 & \textbf{41.2 / 81.2} & 25.3 / 59.6 & 28.1 / 74.5 & 46.7 / 95.0 & \textbf{52.7 / 97.3} & 35.8 / 98.2 & 40.8 / 97.5 & 46.6 / 97.4 & 43.9 / 98.3 & 37.1 / 94.9 & 38.6 / 98.5 & 39.6 / 88.6 \\
        GPT-4o-mini & 28.3 / 65.7 & 33.4 / 71.5 & 24.9 / 53.9 & 22.6 / 61.1 & 44.6 / 83.8 & 41.9 / 89.4 & 30.0 / 81.8 & \textbf{47.6 / 95.0} & 34.2 / 87.2 & 33.0 / 90.8 & 32.5 / 83.5 & 31.8 / 97.0 & 33.7 / 80.1 \\
        GPT-4-Turbo & 30.8 / 50.0 & 27.9 / 59.1 & 21.8 / 44.9 & 26.7 / 58.6 & 47.8 / 90.7 & 46.0 / 90.3 & 38.8 / 92.7 & 43.5 / 95.0 & 34.6 / 90.1 & 36.8 / 90.8 & 32.4 / 92.4 & 32.5 / 95.5 & 35.0 / 79.2 \\
        GPT-4o      & \textbf{48.3 / 79.6} & 38.1 / 79.0 & \textbf{39.9 / 83.1} & 36.6 / 80.9 & \textbf{51.2 / 91.9} & 50.1 / 89.4 & 36.8 / 87.3 & 41.6 / 95.0 & \textbf{47.4 / 94.9} & 42.5 / 95.8 & \textbf{40.3 / 93.7} & 31.4 / 95.5 & \textbf{42.0 / 88.8} \\
        Gemini-2.5-Flash      & 34.9 / 62.7 & 31.8 / 61.8 & 21.1 / 52.8 & 14.7 / 41.4 & 39.9 / 75.3 & 32.6 / 66.4 & 37.2 / 96.4 & 43.0 / 91.7 & 40.7 / 86.9 & 39.5 / 82.4 & 33.7 / 89.9 & 37.9 / 82.6 & 33.9 / 74.2 \\
        Gemini-2.5-Pro      & 34.1 / 60.3 & 29.6 / 62.9 & 26.2 / 69.7 & 24.7 / 63.1 & 35.7 / 66.4 & 25.8 / 54.0 & 22.8 / 58.2 & 11.7 / 23.3 & 37.3 / 68.6 & 25.7 / 65.5 & 26.4 / 69.6 & 25.6 / 57.6 & 27.1 / 59.9 \\
        Grok-3 & 47.2 / 92.3 & 19.1 / 38.2 & 38.0 / 89.9 & \textbf{38.9 / 93.6} & 44.1 / 96.9 & 51.3 / 92.0 & \textbf{39.0 / 96.4} & 39.0 / 95.8 & 41.1 / 96.7 & 39.1 / 95.8 & 37.1 / 96.2 & \textbf{40.9 / 97.0} & 39.6 / 90.1 \\
        Grok-4 & 41.9 / 75.6 & 38.8 / 83.3 & 22.9 / 75.3 & 27.2 / 81.5 & 41.4 / 98.1 & 47.6 / 98.2 & 30.8 / 98.2 & 32.5 / 96.7 & 45.0 / 93.4 & 42.8 / 95.0 & 29.8 / 88.6 & 37.6 / 89.4 & 36.5 / 89.4 \\
        \midrule
        \multicolumn{14}{c}{\textbf{Open-source LLMs (\textasciitilde 7B)}} \\
        \midrule
        DSC-6.7B-Instruct            & 12.8 / 35.4 & 15.1 / 41.4 & 6.3 / 20.2 & 12.6 / 28.0 & 12.8 / 32.4 & 6.9 / 15.9 & 24.6 / 87.3 & 19.2 / 49.2 & 16.0 / 36.9 & 17.2 / 50.4 & 17.4 / 53.2 & 7.7 / 25.8 & 14.0 / 39.7 \\
        Qwen2.5-Coder-7B-Instruct    & 18.2 / 43.7 & 19.5 / 54.8 & 9.2 / 38.2 & 8.8 / 28.0 & 9.9 / 34.7 & 14.8 / 34.5 & 17.4 / 56.4 & 16.4 / 45.0 & 22.7 / 64.2 & 22.4 / 61.3 & 29.0 / 75.9 & 24.1 / 62.9 & 17.7 / 50.0 \\
        Qwen2.5-7B-Instruct          & 22.3 / 48.6 & 23.7 / 55.9 & 16.3 / 41.6 & 17.1 / 51.0 & 25.8 / 64.1 & 25.4 / 59.3 & 27.9 / 78.2 & 25.3 / 78.3 & 22.3 / 62.0 & 19.9 / 62.2 & 15.1 / 48.1 & 19.0 / 53.0 & 21.7 / 58.5 \\
        OmniSQL-7B & 24.0 / 65.9 & 27.5 / 74.6 & 15.8 / 51.7 & 25.4 / 70.1 & 17.4 / 39.4 & 9.1 / 31.9 & 25.3 / 67.3 & 19.6 / 45.8 & 26.6 / 79.9 & 22.1 / 74.8 & 31.4 / 89.9 & 26.4 / 78.0 & 22.5 / 64.1 \\
        OpenCoder-8B-Instruct        & 15.9 / 36.9 & 21.8 / 52.2 & 12.7 / 48.3 & 17.3 / 46.5 & 14.0 / 32.0 & 10.7 / 24.8 & 22.2 / 80.0 & 16.3 / 50.8 & 24.9 / 57.3 & 23.8 / 60.5 & 26.5 / 82.3 & 15.9 / 37.1 & 18.5 / 50.7 \\
        Meta-Llama-3.1-8B-Instruct   & 12.7 / 41.1 & 14.1 / 44.6 & 12.5 / 40.4 & 9.5 / 35.0 & 10.4 / 36.7 & 12.4 / 35.4 & 11.0 / 63.6 & 10.8 / 28.3 & 16.6 / 42.3 & 20.8 / 55.5 & 19.4 / 65.8 & 15.9 / 39.4 & 13.8 / 44.0 \\
        \midrule
        \textbf{\model-7B-LoRA}      & 37.1 / 79.6 & 35.6 / 82.8 & 30.7 / 77.7 & 28.1 / 79.0 & 39.7 / 90.6 & 42.1 / 91.5 & 37.8 / 95.9 & 39.6 / 94.5 & 44.7 / 96.0 & 42.6 / 95.3 & 41.1 / 96.6 & 38.6 / 95.7 & 38.1 / 89.6 \\
        \textbf{\model-7B}           & \textbf{38.3 / 81.0} & \textbf{36.8 / 84.2} & \textbf{31.9 / 79.1} & \textbf{29.3 / 80.4} & \textbf{40.9 / 92.0} & \textbf{43.3 / 92.9} & \textbf{39.0 / 97.2} & \textbf{40.8 / 95.9} & \textbf{45.9 / 97.3} & \textbf{43.8 / 96.6} & \textbf{42.3 / 97.9} & \textbf{39.8 / 97.0} & \textbf{39.3 / 91.0} \\
        \midrule
        \multicolumn{14}{c}{\textbf{Open-source LLMs (14B-32B)}} \\
        \midrule
        Qwen2.5-Coder-14B-Instruct   & 25.2 / 57.0 & 22.7 / 59.3 & 16.3 / 43.9 & 21.7 / 55.9 & 19.4 / 50.0 & 20.0 / 54.2 & 16.4 / 47.8 & 7.1 / 19.7 & 32.1 / 81.2 & 36.4 / 81.0 & 34.4 / 89.7 & 39.7 / 94.6 & 24.3 / 61.2 \\
        Qwen2.5-14B-Instruct         & 30.5 / 60.6 & 30.5 / 62.4 & 15.4 / 42.7 & 23.2 / 61.1 & 36.1 / 82.2 & 31.8 / 80.5 & 37.9 / 90.9 & 36.0 / 84.2 & 37.8 / 82.8 & 31.4 / 88.2 & 31.3 / 92.4 & 35.2 / 90.2 & 31.4 / 76.5 \\
        OmniSQL-14B & 36.6 / 73.0 & 30.7 / 73.7 & 23.7 / 64.0 & 25.3 / 65.0 & 20.9 / 51.7 & 20.4 / 47.8 & 20.4 / 61.8 & 15.5 / 40.8 & 39.0 / 86.9 & 35.8 / 84.9 & 31.7 / 91.1 & 31.1 / 86.4 & 27.6 / 68.9 \\
        StarCoder2-15B-Instruct      & 16.2 / 41.2 & 16.2 / 50.0 & 14.2 / 38.1 & 12.6 / 29.4 & 10.4 / 22.8 & 4.1 / 10.0 & 19.9 / 61.1 & 10.7 / 34.5 & 25.0 / 59.8 & 25.2 / 68.1 & 25.1 / 66.7 & 13.3 / 41.3 & 16.1 / 43.6 \\
        DSC-V2-Lite-In. (16B, MoE)   & 11.9 / 34.9 & 13.4 / 40.0 & 10.0 / 44.9 & 9.8 / 44.1 & 11.5 / 28.9 & 11.8 / 26.5 & 0.0 / 13.5 & 15.6 / 49.1 & 18.6 / 45.3 & 24.0 / 58.5 & 11.2 / 78.5 & 16.0 / 55.8 & 12.8 / 43.3 \\
        Codestral-22B                & 28.4 / 49.1 & 25.9 / 55.4 & 12.7 / 32.6 & 26.6 / 56.1 & 34.9 / 68.3 & 28.5 / 59.3 & 33.1 / 78.2 & \textbf{43.0 / 89.2} & 38.4 / 86.1 & 36.4 / 79.8 & 32.6 / 83.5 & 37.5 / 86.4 & 31.5 / 68.7 \\
        \midrule
        \textbf{\model-14B-LoRA}     & 40.5 / 84.5 & 38.5 / 86.6 & 34.5 / 82.5 & 32.0 / 84.0 & 43.5 / 94.0 & 45.5 / 94.6 & 41.0 / 97.7 & 42.5 / 96.6 & 47.5 / 98.0 & 45.5 / 97.5 & 44.0 / 98.5 & 41.5 / 97.7 & 41.4 / 92.7 \\
        \textbf{\model-14B}          & \textbf{40.8 / 85.0} & \textbf{38.8 / 87.1} & \textbf{34.8 / 83.0} & \textbf{32.3 / 84.5} & \textbf{43.8 / 94.5} & \textbf{45.8 / 95.1} & \textbf{41.3 / 98.1} & 42.8 / 97.1 & \textbf{47.8 / 98.4} & \textbf{45.8 / 97.9} & \textbf{44.3 / 98.9} & \textbf{41.8 / 98.1} & \textbf{41.7 / 93.1} \\
        \midrule
        \multicolumn{14}{c}{\textbf{Open-source LLMs ($\geq$ 32B)}} \\
        \midrule
        Qwen2.5-Coder-32B-Instruct   & 26.9 / 57.8 & 27.6 / 65.8 & 23.5 / 46.1 & 25.5 / 65.2 & 32.0 / 74.3 & 34.3 / 69.9 & 40.7 / 92.7 & 35.8 / 83.2 & 33.7 / 79.9 & 28.3 / 73.1 & 33.9 / 84.8 & 37.5 / 92.4 & 31.6 / 73.8 \\
        Qwen2.5-32B-Instruct         & 37.0 / 64.8 & 35.4 / 66.7 & 20.5 / 50.6 & 28.2 / 65.0 & 39.8 / 75.3 & 36.1 / 77.0 & 21.8 / 72.7 & 35.4 / 84.2 & 40.4 / 88.0 & 36.1 / 87.4 & 34.0 / 91.1 & 31.1 / 75.0 & 33.0 / 74.8 \\
        OmniSQL-32B & 31.2 / 61.7 & 24.6 / 67.2 & 24.6 / 48.3 & 29.3 / 65.6 & 34.3 / 76.1 & 35.3 / 78.8 & 30.9 / 83.6 & 28.4 / 70.8 & 30.7 / 67.9 & 26.3 / 64.7 & 30.6 / 81.0 & 30.0 / 83.3 & 29.7 / 70.8 \\
        DSC-33B-Instruct             & 18.5 / 45.3 & 17.9 / 51.4 & 7.2 / 29.2 & 14.6 / 40.8 & 9.6 / 23.9 & 4.6 / 9.8 & 15.4 / 80.0 & 8.8 / 26.9 & 23.4 / 64.8 & 23.8 / 69.7 & 21.9 / 83.5 & 22.8 / 65.2 & 15.7 / 49.2 \\
        Meta-Llama-3.1-70B-Instruct  & 28.3 / 53.8 & 22.9 / 61.3 & 19.9 / 50.6 & 19.5 / 56.7 & 30.5 / 71.0 & 40.3 / 75.2 & 32.7 / 89.1 & 5.2 / 11.7 & 38.9 / 87.2 & 34.2 / 85.7 & 34.0 / 86.1 & 36.3 / 92.4 & 28.6 / 68.4 \\
        Qwen2.5-72B-Instruct         & 43.4 / 73.5 & 38.0 / 70.4 & 23.9 / 59.6 & 32.0 / 66.2 & 41.4 / 86.1 & 37.8 / 77.9 & 41.1 / 94.5 & 43.6 / 88.3 & 39.4 / 88.0 & 37.8 / 86.6 & 37.8 / 83.5 & 37.4 / 87.9 & 37.8 / 80.2 \\
        DeepSeek-V3.1 (671B, MoE)      & \textbf{43.7 / 84.3} & \textbf{42.3 / 84.9} & 34.4 / 79.8 & 29.4 / 83.4 & \textbf{47.6 / 92.7} & \textbf{50.9 / 92.9} & \textbf{45.4 / 100.0} & 38.6 / 93.3 & 42.6 / 95.6 & 44.6 / 97.5 & 36.1 / 96.2 & 37.9 / 95.5 & 41.1 / 91.3 \\
        \midrule
        \textbf{\model-32B}          & 41.8 / 86.1 & 39.8 / 88.0 & \textbf{35.8 / 84.1} & \textbf{33.3 / 85.3} & 44.8 / 95.5 & 46.8 / 96.0 & 42.3 / 98.5 & \textbf{43.8 / 97.6} & \textbf{48.8 / 98.9} & \textbf{46.8 / 98.4} & \textbf{45.3 / 99.3} & \textbf{42.8 / 98.6} & \textbf{42.7 / 93.9} \\
        \bottomrule
    \end{tabular}
     }
\end{table*}

\subsubsection{Implementation Details}
\label{sec:impl_details}
The UniVectorSQL model is developed by fine-tuning the open-source OmniSQL model, which is itself an NL2SQL model fine-tuned from the Qwen2.5-Coder that trained on 92 programming languages \cite{qwen2.5_coder}. \model-7B and \model-14B are fully fine-tuned, while \model-32B is fine-tuned using a parameter efficient technique, low-rank adaptation (LoRA)~\cite{Hu2022@lora}. The training data comprises our SynVecSQL corpus, with optional SynSQL dataset from OmniSQL to retain strong standard SQL capabilities in data mixing experiments. For LoRA, we set $r = 256$ and $\alpha = 512$. We use the AdamW optimizer~\cite{Loshchilov2019@adamw} with parameters $\beta_{1} = 0.9$, $\beta_{2} = 0.95$, and $\epsilon = 10^{-8}$ to optimize the training objective. The peak learning rates are set to $2e^{-5}$, $4e^{-6}$ and $2e^{-4}$ for \model-7B, \model-14B, and \model-32B, respectively. We use a learning rate schedule with a linear warmup for the initial 5\% of training, followed by cosine decay to 10\% of the peak rate. In addition, the batch size, number of epochs (with early-stop), context length, weight decay, and gradient clipping are uniformly set to 64, 2, 8192, 0.1, and 1.0, respectively. To optimize GPU memory usage during training, we leverage the DeepSpeed with bfloat16 mixed precision and ZeRO3 optimization~\cite{Rajbhandari2020@deepspeed}. In practice, training \model-7B/14B/32B finishes in one day on a single machine equipped with 4 NVIDIA A100-SXM4-80GB GPUs. We also provide LoRA versions of \model-7B and \model-14B (referred to as \model-7B-LoRA and \model-14B-LoRA), using the same LoRA hyperparameters as \model-32B. This enables a direct comparison between full and LoRA-based fine-tuning approaches. For embedding models, we employ all-MiniLM-L6-v2 \cite{reimers-2019-sentence-bert} on BIRD, Spider, and arXiv benchmarks and during training data synthesis. For Wikipedia, we adopt CLIP-ViT-B-32-laion2B \cite{Radford2021LearningTV}. The backends for evaluation are SQLite 3.46 (with SQLite-vec 0.1.6), PostgreSQL 14.18, and ClickHouse 25.9. For a realistic evaluation, we kept default execution settings but set a 60s execution timeout and built HNSW indices on ClickHouse and PostgreSQL to accelerate vector search and query evaluation.

\subsection{Overall Performance}
Table \ref{tab:backend_results} details the main evaluation results, comparing our \model{} models against a wide array of open-source and proprietary LLMs. Our models, fine-tuned on SynVecSQL-24K, establish a dominant state-of-the-art performance among all open-source competitors. \model-7B, with an average nDCG@10 of 39.3 and a 91.8\% success rate, drastically outperforms other models in its class, such as OmniSQL-7B (22.5 nDCG@10) and Qwen2.5-Coder-7B (17.7 nDCG@10). This performance also scales, with \model-32B achieving a 42.7 nDCG@10, surpassing even the much larger DeepSeek-V3.1 (41.1) and Qwen2.5-72B (37.8).

When benchmarked against leading closed-source APIs, our \model-32B model proves to be highly competitive (42.7) and slightly edges out top-tier models like GPT-4o (42.0), Grok-3 (39.6), and Claude-4-Sonnet (39.6). Crucially, \model-32B also demonstrates a superior execution success rate (93.9\%) compared to GPT-4o (88.8\%), highlighting its strong reliability in generating syntactically correct queries across all three database backends.

The results also reveal the distinct challenge of the Text2VectorSQL task, as even strong Text2SQL models like OmniSQL and DeepSeek-Coder show poor performance without specialized fine-tuning. 

\subsection{Ablation Studies and Discussions}
\label{sec:ablation}
To validate our key design choices and provide deeper insights, we conduct a series of ablation studies and discussions. We investigate four central aspects: the impact of the training data mixing, the efficacy of chain-of-thought, the recall degradation phenomenon and intuitive case studies.

\paragraph{Impact of Training Data Mixture}
\label{subsec:ablation_data}

Table~\ref{tab:ablation_data} presents the ablation study on training data composition. We use the OmniSQL-7B model, originally trained on SynSQL, as the starting point for this comparison. The base OmniSQL model performs reasonably well on the standard BIRD benchmark (63.9\% EX) but fails at the new Text2VectorSQL task, achieving only 22.5\% nDCG@k on VectorSQLBench. Its decomposed accuracy shows that while it can sometimes generate a valid SQL skeleton (83.2\% ACC\textsubscript{SQL}), it is incapable of formulating the vector component (41.3\% ACC\textsubscript{Vec}).

The model trained solely on our SynVecSQL data (1:0 ratio) achieves the highest performance on VectorSQLBench (39.3 nDCG@k), demonstrating its effectiveness for the target task. However, this comes at the cost of degradation on standard Text-to-SQL (58.2 EX on BIRD-dev), indicating noticeable catastrophic forgetting of previously acquired SQL skills.

Introducing SynSQL data into the training mixture effectively mitigates this issue. As the proportion of SynSQL increases, performance on BIRD-dev improves, reaching 63.2 EX at a 1:4 ratio. However, this improvement is also accompanied by a corresponding decline in VectorSQLBench performance (32.7 nDCG@k), highlighting the importance of high-quality data preparation and mixing.

The 1:1 mixing ratio strikes a satisfactory balance, achieving strong performance on both BIRD-dev (61.5 EX) and VectorSQLBench (37.8 nDCG@k). This balanced approach ensures the model acquires new vector search capabilities while retaining its mastery over complex traditional SQL, validating our blended training strategy as essential for building versatile unified query interfaces. Given our focus on establishing a robust baseline for the Text2VectorSQL task, we opted for the pure SynVecSQL (1:0) ratio for our \model{} models.

\begin{table}[t]
\centering
\caption{Ablation study on training data composition on \model-7B. We evaluate on BIRD-dev (Execution Accuracy) and VectorSQLBench (nDCG@k and Decomposed Accuracy). The mixture yields the best overall performance, demonstrating the necessity of a blended training strategy.}
\label{tab:ablation_data}
\renewcommand{\arraystretch}{1.15}
\resizebox{\columnwidth}{!}{%
\begin{tabular}{@{}ccccc@{}}
\toprule
\multirow{3}{*}{\textbf{\makecell[c]{Mixing Ratio \\ {\scriptsize (SynVecSQL:SynSQL)}}}} & \multicolumn{1}{c}{\textbf{BIRD-dev}} & \multicolumn{3}{c}{\textbf{VectorSQLBench}} \\
\cmidrule(lr){2-2} \cmidrule(lr){3-5}
& EX (\%) & nDCG@k (\%) & ACC\textsubscript{SQL} (\%) & ACC\textsubscript{Vec} (\%) \\
\midrule
OmniSQL-7B & 63.9 & 22.5 & 83.2 & 41.3 \\
\midrule
1:0 & 58.2 & 39.3 & 89.7 & 78.5 \\
1:1 & 61.5 & 37.8 & 88.3 & 76.2 \\
1:2 & 62.8 & 35.4 & 87.1 & 72.9 \\
2:1 & 60.1 & 38.5 & 89.1 & 77.8 \\
1:4 & 63.2 & 32.7 & 85.9 & 68.4 \\
4:1 & 58.3 & 39.1 & 89.4 & 78.1 \\
\bottomrule
\end{tabular}
}
\end{table}

\paragraph{Impact of Chain-of-Thought}

\begin{table*}[t]
    \centering
    \caption{Ablation study on the impact of Chain-of-Thought (CoT). Performance evaluation (nDCG@10) and execution success rate (in \%) of UniVectorSQL with and without CoT on VectorSQLBench across three database backends. Our CoT introduces consistent and considerable performance uplift across all model scales, especially for small-scale models.
    }
    \label{tab:ablation_cot_detailed}
    \setlength{\tabcolsep}{3pt}
    \renewcommand{\arraystretch}{1.1}
    \resizebox{\textwidth}{!}{
    \begin{tabular}{l|l|cccc|cccc|cccc|c}
        \toprule
        \multirow{2.5}{*}{\textbf{Model}} & \multirow{2.5}{*}{\textbf{Setting}} & \multicolumn{4}{c|}{\textbf{SQLite}} & \multicolumn{4}{c|}{\textbf{PostgreSQL}} & \multicolumn{4}{c|}{\textbf{ClickHouse}} & \multirow{2.5}{*}{\textbf{Average}} \\
        \cmidrule(lr){3-6} \cmidrule(lr){7-10} \cmidrule(lr){11-14}
        & & \textbf{BIRD} & \textbf{Spider} & \textbf{arXiv} & \textbf{Wiki} & \textbf{BIRD} & \textbf{Spider} & \textbf{arXiv} & \textbf{Wiki} & \textbf{BIRD} & \textbf{Spider} & \textbf{arXiv} & \textbf{Wiki} & \\
        \midrule
        \multirow{2}{*}{\model-7B} & w/o CoT & 33.9 / 74.6 & 32.4 / 77.8 & 27.5 / 72.7 & 24.9 / 74.0 & 36.5 / 85.6 & 38.9 / 86.5 & 34.6 / 90.8 & 36.4 / 89.5 & 41.5 / 90.9 & 39.4 / 90.2 & 37.9 / 91.5 & 35.4 / 90.6 & 34.9 / 84.6 \\
        & w/ CoT & \textbf{38.3 / 81.0} & \textbf{36.8 / 84.2} & \textbf{31.9 / 79.1} & \textbf{29.3 / 80.4} & \textbf{40.9 / 92.0} & \textbf{43.3 / 92.9} & \textbf{39.0 / 97.2} & \textbf{40.8 / 95.9} & \textbf{45.9 / 97.3} & \textbf{43.8 / 96.6} & \textbf{42.3 / 97.9} & \textbf{39.8 / 97.0} & \textbf{39.3 / 91.0} \\
        \midrule
        \multirow{2}{*}{\model-14B} & w/o CoT & 36.6 / 78.8 & 34.6 / 80.9 & 30.6 / 76.8 & 28.1 / 78.3 & 39.6 / 88.3 & 41.6 / 88.9 & 37.1 / 91.9 & 38.6 / 90.9 & 43.6 / 92.2 & 41.6 / 91.7 & 40.1 / 92.7 & 37.6 / 91.9 & 37.5 / 86.9 \\
        & w/ CoT & \textbf{40.8 / 85.0} & \textbf{38.8 / 87.1} & \textbf{34.8 / 83.0} & \textbf{32.3 / 84.5} & \textbf{43.8 / 94.5} & \textbf{45.8 / 95.1} & \textbf{41.3 / 98.1} & \textbf{42.8 / 97.1} & \textbf{47.8 / 98.4} & \textbf{45.8 / 97.9} & \textbf{44.3 / 98.9} & \textbf{41.8 / 98.1} & \textbf{41.7 / 93.1} \\
        \midrule
        \multirow{2}{*}{\model-32B} & w/o CoT & 38.6 / 81.9 & 36.6 / 83.8 & 32.6 / 79.9 & 30.1 / 81.1 & 41.6 / 91.3 & 43.6 / 91.8 & 39.1 / 94.3 & 40.6 / 93.4 & 45.6 / 94.7 & 43.6 / 94.2 & 42.1 / 95.1 & 39.6 / 94.4 & 39.5 / 89.7 \\
        & w/ CoT & \textbf{41.8 / 86.1} & \textbf{39.8 / 88.0} & \textbf{35.8 / 84.1} & \textbf{33.3 / 85.3} & \textbf{44.8 / 95.5} & \textbf{46.8 / 96.0} & \textbf{42.3 / 98.5} & \textbf{43.8 / 97.6} & \textbf{48.8 / 98.9} & \textbf{46.8 / 98.4} & \textbf{45.3 / 99.3} & \textbf{42.8 / 98.6} & \textbf{42.7 / 93.9} \\
        \bottomrule
    \end{tabular}
    }
\end{table*}

To quantify the benefit of our structured reasoning approach, we conducted an ablation study on the inclusion of Chain-of-Thought (CoT) during fine-tuning. We trained two versions of each \model{} model: one using the full training sequence including the CoT (`w/ CoT'), and another trained to directly map the input to the final VectorSQL query (`w/o CoT'). The results, detailed in Table~\ref{tab:ablation_cot_detailed}, demonstrate a consistent performance uplift across all model scales when CoT is utilized.

For instance, the \model-32B model with CoT achieved an average nDCG@10 of 42.7, a 3.2-point improvement over the 39.5 nDCG@10 from the counterpart trained without CoT. This performance gain was mirrored in query reliability, as the execution success rate also rose from 89.7\% to 93.9\%. Similar gains were observed for the 7B and 14B models, confirming that training the model to explicitly articulate its reasoning process is crucial for mastering the complex, hybrid nature of the Text2VectorSQL task.

This improvement can be attributed to the CoT's role as a cognitive scaffold. Text2VectorSQL is uniquely complex as it requires the model to simultaneously manage two distinct reasoning paths: the structured path for SQL predicates on relational data and the semantic path for vector search over unstructured data. Our specialized CoT (as described in Section \ref{subsubsec:step4_cot}) forces the model to first decompose this hybrid intent, plan the SQL and vector components separately, and then reason about their final assembly. This step-by-step process significantly reduces the task's complexity, helping the model avoid syntactic errors and logical fallacies, thereby generating more accurate and robust queries.

\paragraph{The Recall Degradation Phenomenon}
A critical challenge in hybrid query execution, which has been largely unexamined in the Text2VectorSQL context, is \textbf{recall degradation}. In practice, both PostgreSQL and ClickHouse adopt post-filtering for efficiency, but also at a significant risk: the true top-$k$ results that satisfy the SQL predicates can be absent from vector search results. Our analysis reveals that this phenomenon becomes pronounced for complex VectorSQL, particularly with queries involving \texttt{JOIN} operations. That is, a \texttt{JOIN} clause can act as a highly selective and data-dependent filter that is impossible for a standard ANN index to anticipate.

\begin{figure*}[tb]
    \centering
    \includegraphics[width=1.0\linewidth]{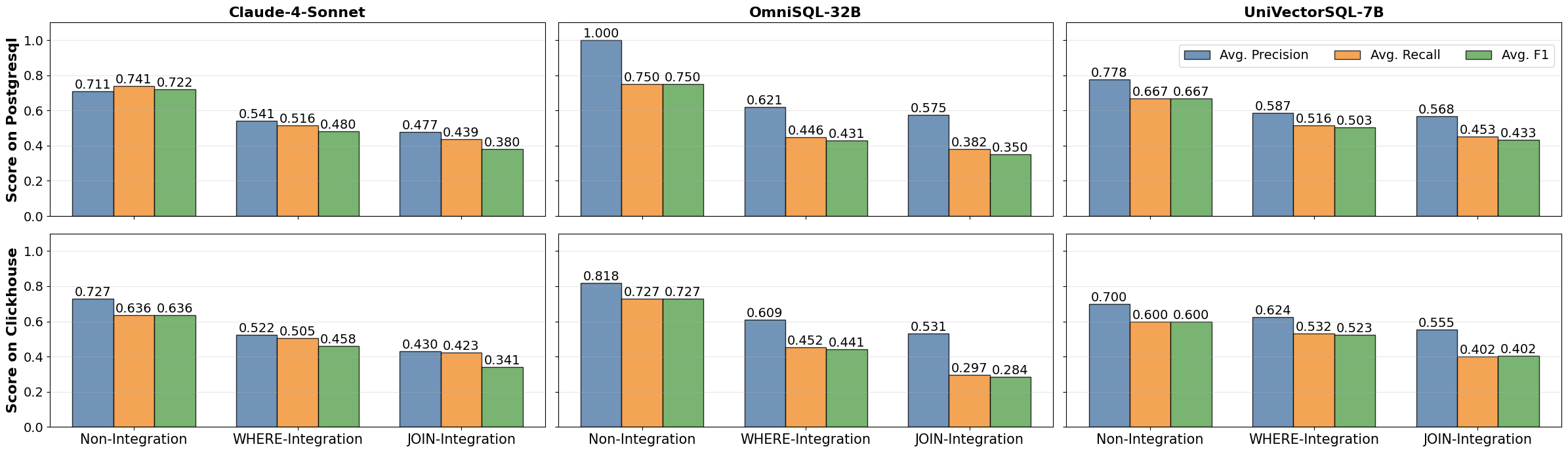}
    \caption{Recall degradation phenomenon across different models on PostgreSQL and ClickHouse. The charts show Avg. Precision, Avg. Recall, and Avg. F1 of four datasets in VectorSQLBench, categorized by hybrid integration depth. A severe drop in Avg. Recall (orange bar) is observed as the integration complexity increases from \textit{Non-integration} to \textit{WHERE-Integration} and is most pronounced in \textit{JOIN-Integration} queries, highlighting a critical challenge in Text2VectorSQL execution.}
    \label{fig:recall_degradation}
\end{figure*}

To empirically quantify this problem, we evaluated three representative models (Claude-4-Sonnet, OmniSQL-32B, and our UniVectorSQL-7B) on the PostgreSQL and ClickHouse backends, with results segmented by the vectorial integration complexity defined in Section \ref{subsubsec:step2_synthesis}.

Figure \ref{fig:recall_degradation} provides clear evidence. Across all models, there is a significant drop in average recall (orange bars) as the query complexity moves from simple \textit{Non-integration} to \textit{WHERE-Integration}, and it plummets in the most complex \textit{JOIN-Integration} scenarios. For instance, on ClickHouse, UniVectorSQL-7B's recall degrades by more than 30\% (from Non-integration's 0.6 to JOIN-integration's 0.4). However, while PostgreSQL also suffers from this problem, its degradation is much less severe. In fact, pgvector has introduced iterative index scans \cite{ni2023diskann++}. This mechanism automatically scans more of the index until enough results are found, providing a remedy to the recall degradation caused by post-filtering.

The comparison of recall degradation between ClickHouse and PostgreSQL isolates the inherent problem complexity factor, and highlights that: the post-filtering execution strategy itself, especially when combined with complex SQL joins, can fail to retrieve a large portion of correct results. This underscores that Text2VectorSQL is not just a translation problem; it is deeply linked to the underlying database's execution strategy and query optimization capabilities.

\paragraph{Case Studies}
To provide a more intuitive understanding of the challenges in Text2VectorSQL and the capabilities of our model, Table~\ref{tab:case_study} presents a comparative case study. We analyze the generated queries from our \model-7B and a powerful open-source baseline, Qwen2.5-Coder-7B, for a typical hybrid user question. This example highlights a common and subtle failure mode where models misinterpret the boundary between structured filters and semantic search topics, leading to logically flawed queries.

\begin{table}[h!]
\centering
\caption{A case study comparing query generation for a complex hybrid question. \model{} correctly separates structured and semantic constraints, while the baseline conflates them, leading to an incorrect query.}
\label{tab:case_study}
\renewcommand{\arraystretch}{1.3}
\resizebox{\columnwidth}{!}{%
\begin{tabular}{p{0.2\columnwidth}|p{0.8\columnwidth}}
\toprule
\multicolumn{2}{p{\columnwidth}}{\textbf{Natural Language Question}: "Find the 5 papers from the ICDE conference published after 2022 that are most related to query optimization techniques."} \\
\midrule
\makecell[c]{\\\textbf{\model} \\\textbf{-7B}} & \texttt{SELECT title, conference, publication\_year FROM papers WHERE conference = 'ICDE' AND publication\_year > 2022 ORDER BY abstract\_embedding <-> EMBED('query optimization techniques') LIMIT 5;} \\
\midrule
\multicolumn{2}{p{\columnwidth}}{\textbf{Analysis}: \textbf{Correct}. The model successfully decomposes the hybrid intent. It correctly separates the structured predicates (\texttt{conference = 'ICDE'}, \texttt{publication\_year > 2022}) from the semantic search topic (\texttt{'query optimization techniques'}) and assembles them into a logically sound and complete query.} \\
\midrule
\makecell[c]{\\\textbf{Qwen2.5-Coder}\\\textbf{-7B}} & \texttt{SELECT title, conference, publication\_year FROM papers WHERE publication\_year > 2022 ORDER BY abstract\_embedding <-> EMBED('query optimization techniques on ICDE') LIMIT 5;} \\
\midrule
\multicolumn{2}{p{\columnwidth}}{\textbf{Analysis}: \textbf{Incorrect}. While the baseline correctly handles the date filter, it fails to distinguish between a structured filter and a semantic concept. It incorrectly conflates the ICDE conference constraint with the search topic, embedding ``query optimization techniques on ICDE". This fundamentally changes the semantic query and, more importantly, omits the required hard filter (\texttt{WHERE conference = 'ICDE'}), leading to logically incorrect results.} \\
\bottomrule
\end{tabular}%
}
\end{table}

\section{Related Work} \label{sec:related_work}
\textbf{Relational and Vector Databases.}
The data management landscape has historically been dominated by relational databases, which excel at storing and querying structured data using SQL. However, with the explosive growth of unstructured data, projected to constitute 80\% or even 92.9\% of all data \cite{king80percent,idc92}, a new paradigm has emerged. Vector databases have become pivotal for managing high-dimensional embeddings derived from unstructured sources like text and images. Foundational technologies such as Locality Sensitive Hashing (LSH) \cite{charikar2002similarity}, product quantization \cite{jegou2010product}, and graph-based indexing methods like HNSW \cite{malkov2018efficient} have enabled efficient Approximate Nearest Neighbor (ANN) search. This has led to the development of powerful open-source libraries like Faiss \cite{johnson2019billion} and vector databases such as Milvus \cite{wang2021milvus}, Pinecone \cite{bruch2023approximate} and MyScale \cite{myscale}. Despite these advancements, the relational and vector paradigms have largely operated in silos. While some systems are beginning to integrate vector operations, they typically rely on manual crafting of hybrid queries. This leaves a significant gap in creating a unified, user-friendly interface. Our task, Text2VectorSQL, is the first to bridge this divide at the natural language interaction layer.

\textbf{Text-to-SQL Frameworks and Benchmarks.}
The task of translating natural language into SQL has evolved significantly, from early encoder-decoder architectures enhanced with graph neural networks to capture database schema relations \cite{Wang2020@ratsql, Cao2021@lgesql}, to fine-tuning large pre-trained models like T5 \cite{Li2023@resdsql, Scholak2021@picard}, and most recently, sophisticated agentic frameworks based on LLMs \cite{Pourreza2023@dinsql, Gao2024@dailsql, Wang2025@macsql, fan2025grounding,fan2024metasql,fan2023gensql,chronis2025filtered}, which decomposes the Text-to-SQL task into sub-problems like schema linking, query generation, and self-correction. This progress has been heavily driven and measured by a series of high-quality benchmarks. Seminal datasets like Spider \cite{yu2018spider} established a cross-domain standard, while subsequent benchmarks introduced new complexities, such as execution efficiency in BIRD \cite{li2024bird}, database documentation in KaggleDBQA \cite{lee2021kaggledbqa}, and intricate domain-specific queries in MIMICSQL \cite{wang2020MIMICSQL}. However, a common limitation unites these advancements: both the frameworks and the benchmarks operate exclusively on structured, relational data. They cannot interpret or generate queries involving semantic similarity or cross-modal matching, which are central to modern data applications. This absence makes it impossible to evaluate the ability to perform combined structured and unstructured queries. To fill this void, we introduce VectorSQLBench as the first benchmark to provide the necessary infrastructure to measure and drive progress in this new, unified query paradigm, and UniVectorSQL to enable automatic VectorSQL translation. This moves beyond merely improving SQL generation accuracy to fundamentally broadening the scope of questions that can be answered.

\textbf{Filtered Vector Search.}
Filtered vector search, which combines ANN search with attribute-based predicates, is a well-known simplified form of VectorSQL. The two canonical strategies are pre-filtering and post-filtering \cite{vbase, milvus2022hybridsearch}. Pre-filtering first narrows down the dataset based on the structured predicate and performs an ANN search on the filtered subset. Conversely, post-filtering retrieves candidate vectors and then filters out those that do not satisfy the predicate. While straightforward and efficient, post-filtering is prone to low recall, as the true neighbors satisfying the predicate may not be among the initially retrieved candidates, a challenge that becomes severe when filters are highly selective \cite{filtereddiskANN}. To overcome limitations, a significant body of research has focused on creating dedicated index structures that jointly consider vector and attribute information. These include adaptations of graph-based indexes like DiskANN \cite{jayaram2019diskann, gollapudi2023filtered}, partitioning-based methods that cluster data by attributes \cite{gupta2023caps, mohoney2023high}, and composite indexes that integrate attributes directly into vector index structure \cite{wang2022navigable, acorn}. Although these specialized indexes improve efficiency and recall over naive approaches, they require low-level programmatic interaction \cite{li2025sieve,engels2024approximate} and may not apply to complex VectorSQL scenarios. Our work is the first to investigate these challenges at the SQL abstraction layer, revealing that the complexity of VectorSQL can exacerbate the post-filtering recall issue, a critical problem that has been overlooked in the pursuit of a unified query interface.

\section{Conclusion}
\label{sec:conclusion}
In this work, we introduce and formalize Text2VectorSQL, a novel task for unifying natural language queries over structured and unstructured data. To catalyze research, we developed a foundational ecosystem featuring VectorSQLGen, a scalable data synthesis pipeline, and VectorSQLBench, the first comprehensive benchmark for the task. Our resulting \model{} models establish a new state-of-the-art as open-source solutions. Critically, our analysis uncovers the challenge of recall degradation, where integrating complex SQL filters like \texttt{JOIN}s with vector search leads to silent omissions of correct results. This finding elevates Text2VectorSQL from a simple translation task to a complex query co-optimization problem. Our work lays the essential groundwork—defining the task, providing tools, and identifying key challenges—for building the next generation of unified data interfaces.

\section*{Acknowledgement}
This work is supported by the National Key R\&D Program of China (2024YFA1014003), National Natural Science Foundation of China (92470121, 62402016), CAAI-Ant Group Research Fund, and High-performance Computing Platform of Peking University.

\newpage

\section*{AI-Generated Content Acknowledgement}
We acknowledge the use of the generative AI assistant Gemini-2.5-Pro \cite{comanici2025gemini} in the preparation of this article. This tool was applied throughout the manuscript (Sections \ref{sec:introduction} through \ref{sec:conclusion}) to improve language, grammar, word choice, and overall readability. The level of use was strictly limited to text polishing and refinement. All core scientific contributions, including the formalization of the Text2VectorSQL task, the design of the VectorSQLGen pipeline and VectorSQLBench, the experimental methodology, and the analysis of results, were developed exclusively by the human authors. The authors meticulously reviewed and edited all AI-suggested modifications to ensure the final content is accurate and reflects their original work.

\bibliographystyle{IEEEtran}
\bibliography{main.bib}

\end{document}